\documentclass[conference]{IEEEtran}
\IEEEoverridecommandlockouts
\usepackage{cite}
\usepackage{amsmath,amssymb,amsfonts}
\usepackage{graphicx}
\usepackage{textcomp}
\usepackage{xcolor}
\usepackage{multirow}
\usepackage{arydshln}
\usepackage{todonotes}
\usepackage[backref, colorlinks,urlcolor=blue]{hyperref}
\usepackage{etoolbox}
\usepackage{setspace}
\usepackage[bottom]{footmisc}
\usepackage{placeins}
\usepackage{wrapfig}
\usepackage[export]{adjustbox}
\usepackage{amsmath}
\usepackage{algpseudocode}
\usepackage[ruled,vlined]{algorithm2e}
\usepackage{tikz}
\usepackage{relsize}
\usepackage{mathtools, nccmath}
\def\BibTeX{{\rm B\kern-.05em{\sc i\kern-.025em b}\kern-.08em
    T\kern-.1667em\lower.7ex\hbox{E}\kern-.125emX}}
    
\usepackage{url}

\newcommand{\fig}{Figure }
\patchcmd{\section}{\centering}{}{}{}
\newcommand{\greencheck}{}%
\DeclareRobustCommand{\greencheck}{%
  \tikz\fill[scale=0.4, color=green]
  (0,.35) -- (.25,0) -- (1,.7) -- (.25,.15) -- cycle;%
}

\begin{document}
    
\title{
   Experimental Exploration of Compact Convolutional Neural Network Architectures for Non-temporal Real-time Fire Detection
    \vspace{-0.3cm}
}

 \author{\IEEEauthorblockN{Ganesh Samarth C.A.$^{1,2}$,
   Neelanjan Bhowmik$^{2}$,
     Toby P. Breckon$^{2,3}$
   }
 \IEEEauthorblockA{$^{1}$Department of Electrical Engineering, Indian Institute of Technology Dharwad, India \\
 Department of \{Computer Science$^2$ $|$ Engineering$^3$\}, Durham University, UK}
 }
\maketitle
\begin{abstract}
 In this work we explore different Convolutional Neural Network (CNN) architectures and their variants for non-temporal binary fire detection and localization in video or still imagery. We consider the performance of experimentally defined, reduced complexity deep CNN architectures for this task and evaluate the effects of different optimization and normalization techniques applied to different CNN architectures (spanning the Inception, ResNet and EfficientNet architectural concepts). Contrary to contemporary trends in the field, our work illustrates a maximum overall accuracy of 0.96 for full frame binary fire detection and 0.94 for superpixel localization using an experimentally defined reduced CNN architecture based on the concept of InceptionV4. 
We notably achieve a lower false positive rate of 0.06 compared to prior work in the field presenting an efficient, robust and real-time solution for fire region detection.
\end{abstract}


\section{Introduction} \label{sec:intro}
Automated fire detection and localization have become essential tasks in the modern day auto-monitoring systems. The increasing prevalence of industrial, public space and general environment monitoring using security-driven CCTV video systems has given rise to the consideration of these systems as sources of initial fire detection. Furthermore, the on-going consideration of remote vehicles for fire detection and monitoring tasks \cite{bardshaw91robots,MARTINEZDEDIOS2006uavfire,zhang09fire} further enhances the demand for autonomous fire detection from such platforms. Fire detection stands out among other object classification tasks as fire does not have a definite shape or pattern but instead varies with the underlying material composition.

Most early work revolves around a color, texture and shape based approach to fire detection and localization. A color threshold based approach is explored in \cite{Healey93fire} which is extended with the basic consideration of motion by \cite{Phillips02fire}. Later work considers the temporal variation of fire in the Fourier domain \cite{liu2004fire} with progressive studies formulating the problem as a Hidden Markov Model \cite{Toreyin2005fire}. 
More recent works consider machine learning based classification approaches to the fire detection problem \cite{zhang09fire,Ko2009fire,Chen11:Fire}. Chenebert  et al. \cite{Chen11:Fire} consider the use of a non-temporal approach along with colour and texture feature descriptions as an input to a shallow neural network classifier.

With the advent of deep learning, further developments to fire detection based on deep CNN architectures now perform binary fire detection more efficiently and robustly compared to earlier color based approaches \cite{Healey93fire,chen04fire}. 
Attempts to detect fire based on robust smoke detection \cite{xu2017deep} have been made by using synthetically produced smoke images. Some recent work \cite{shen2018flame} applies YOLO architecture \cite{redmon2016you} to perform flame detection. There has also been attempt to develop custom architectures \cite{namozov2018efficient} involving convolution, fully connected and pooling layers on a custom dataset created using Generative Adverserial Networks (GAN). 
The work of \cite{sharma2017deep} considers deep CNN architectures such as VGG16 \cite{simonyan2014vgg} and ResNet50 \cite{he2016deep} for the fire detection task. 
A further experimental approach to fire detection \cite{muhammad2018convolutional} is based on exploring InceptionV1 \cite{szegedy2015going} and AlexNet \cite{krizhevsky2012alexnet} architectures.
\begin{figure}[!ht]
\includegraphics[width=\linewidth]{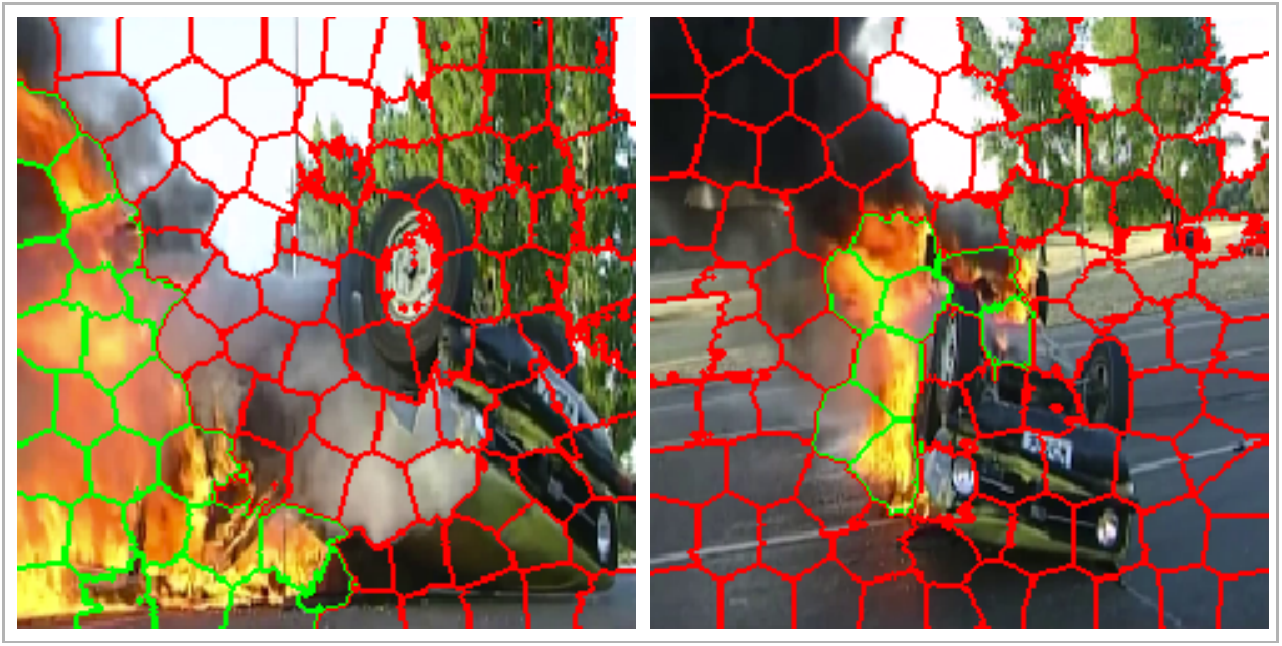}
\vspace{-0.6cm}
\caption{Example fire detection and localization using superpixel (fire = green, no-fire = red).}
\label{fig:examples_fire_local}
\end{figure}

\begin{figure*}[!ht]
\centering
\includegraphics[width=\linewidth]{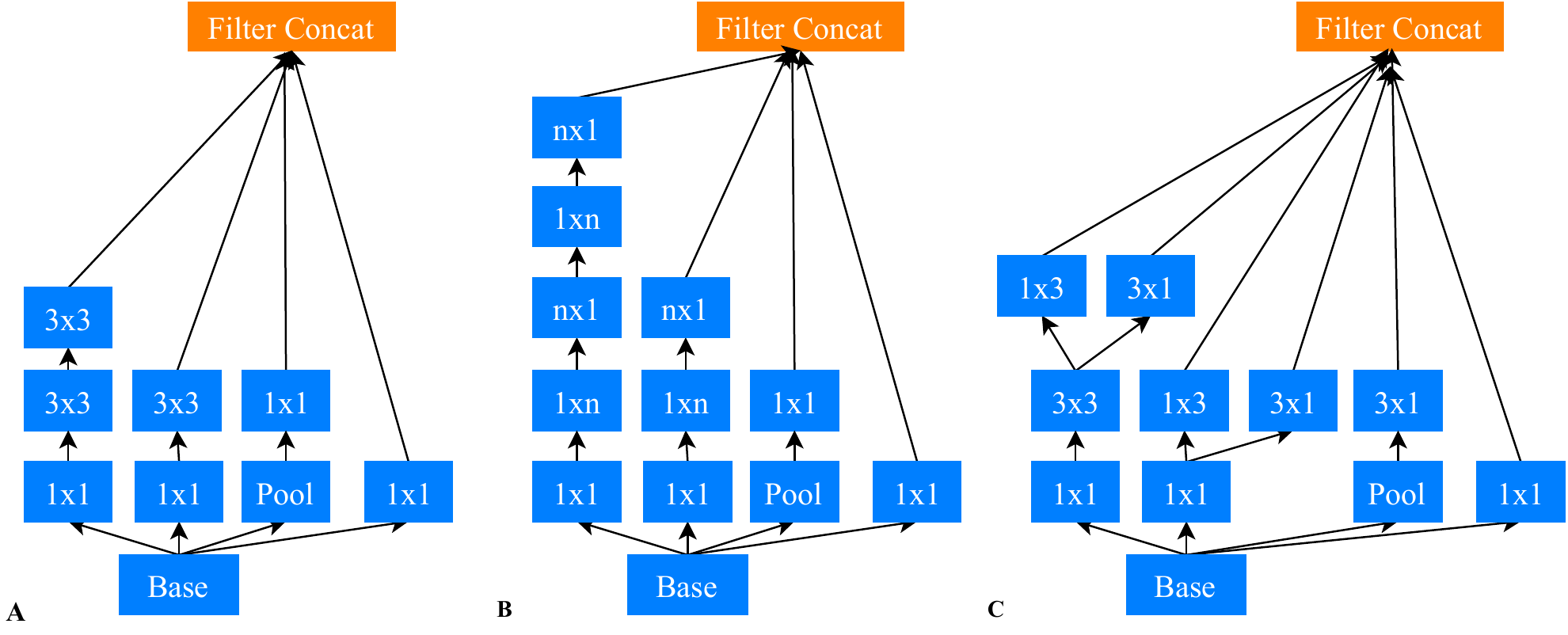}
\vspace{-0.4cm}
\caption{Three variants of InceptionV2 \cite{szegedy2016rethinking}
: Module-A with $3 \times 3$ kernels (A), Module-B with asymmetric convolutions with $n=7$ (B), and Module-C with the wide filter banks of $1 \times 3$ and $3 \times 1$ kernels (C).}
\label{fig:inception_blocks}
\end{figure*}

The work of \cite{dunnings2018experimentally}, a direct precursor to this study, explored fire detection and localization based on both full-frame binary fire detection and superpixels (\fig \ref{fig:examples_fire_local}) based on similar experimentally defined CNN architectural variants derived from the seminal InceptionV1 \cite{szegedy2015going} and AlexNet \cite{krizhevsky2012alexnet} architectures (InceptionV1-OnFire, FireNet, \cite{dunnings2018experimentally}). InceptionV1-OnFire achieved 0.89 detection accuracy for superpixel based detection whilst FireNet achieves 0.92 for full frame binary fire detection.

In this work we expand upon the study of \cite{dunnings2018experimentally} considering a similar experimental approach to the definition of reduced complexity CNN but based on contemporary advances in the Inception  \cite{szegedy2016rethinking}, \cite{szegedy2017inception} and ResNet \cite{he2016deep}, \cite{szegedy2017inception} architectures.
This is a non-temporal approach to the fire detection problem which is highly suited for non-stationary fire detection scenarios. We explicitly consider two fire detection problems: (a) binary fire detection to determine if fire is present in a particular frame  and (b) in-frame superpixel localization to determine the precise location of fire within that frame.

    \section{Proposed Approach}  \label{sec:proposal}
Our approach experimentally defines CNN architectures with low complexity to address the fire detection tasks identified.
Whilst prior work \cite{dunnings2018experimentally} focuses only on AlexNet and InceptionV1 variants, here we expand this remit to additional reference CNN architectures.

\subsection{Reference Architectures} \label{ssec:refArch}
\noindent
\textbf{InceptionV2} \cite{szegedy2016rethinking} is inspired from InceptionV1 \cite{szegedy2015going}. The intuition is that, reducing the dimensions drastically may cause loss of information, known as a ``representational bottleneck". Smart factorization techniques are used to make convolutions more efficient in terms of computational complexity. Hence three different variants of the inception modules are defined (Module - A, B, C). In Module-A, the filters with a $5\times5$ kernel size are replaced with two $3\times3$ kernels (\fig \ref{fig:inception_blocks}A). A $5\times5$ convolution is 2.78 times more expensive than a $3\times3$ convolution. Hence, two $3\times3$ convolutions are connected which leads to a boost in performance. Moreover $ n\times n$  convolutions can be further factorized to a combination of $1\times n$ and $n\times 1$ convolutions. This is found to be three times more efficient and is implemented as the Module-B of the inception variant with $n=7$ (\fig \ref{fig:inception_blocks}B). Filter banks are further expanded to remove the representational bottleneck. The third variant of the inception module, i.e. Module-C (\fig \ref{fig:inception_blocks}C), has $3\times3$ filters factorized into parallel $1\times3$ and $3\times1$ filters and merged to make the module wider. The stem of the network consists of two convolution layers followed by a pooling layer and by three convolution layers. The inception modules are connected with three modules of Module-A, five of Module-B and two of Module-C. This is followed by global max pooling, a linear and a final softmax layer. \\
\begin{figure}[tb]
\centering
\includegraphics[width=6cm]{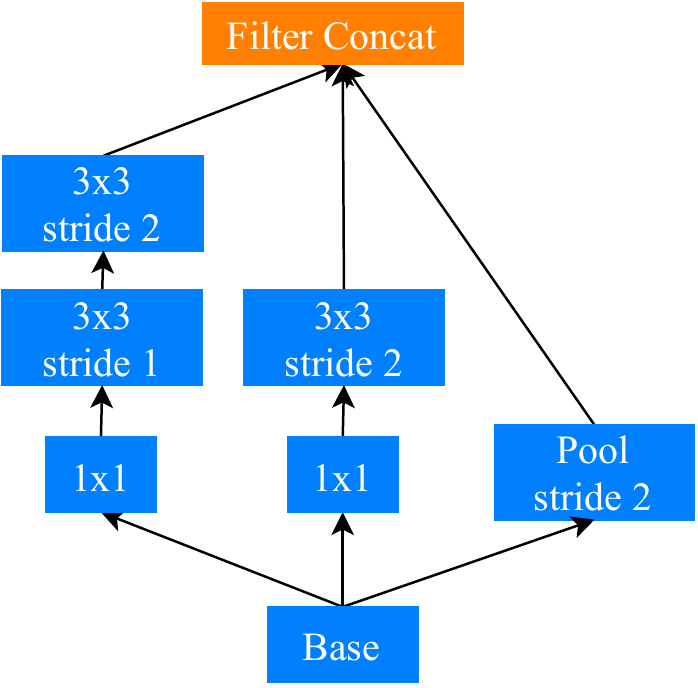}
\vspace{-0.4cm}
\caption{Grid size reduction by connecting convolution layer and max-pooling in parallel in InceptionV3 \cite{szegedy2016rethinking}.}
\label{fig:grid_size_reduction}
\end{figure}
\textbf{InceptionV3} \cite{szegedy2016rethinking} is a very similar architecture to InceptionV2 and it uses same three modular components (\fig \ref{fig:inception_blocks}) additional features, such as Root Mean Square Propagation (RMSProp), label smoothing and batch normalization. Grid size reduction (\fig \ref{fig:grid_size_reduction}) is introduced in this model, where feature maps are connected with a convolution layer of stride two and a max-pooling layer in parallel for concatenation. This is shown better when compared to just using pooling to reduce the dimensions as this leads to some loss of information. The first and second variants of the inception modules have a reduction block succeeding them. \\
\textbf{InceptionV4} \cite{szegedy2017inception} is the next version of Inception proposed (with three variants Module - A,B,C), which differs from its previous version only with respect to its stem. InceptionV4 utilizes the idea of efficient grid size reduction (\fig \ref{fig:grid_size_reduction}) in its stem to reduce the dimensions of the image without any significant loss of information before being passed to the inception and its reduction blocks. \\
\textbf{ResNet} \cite{he2016deep} and variants show very promising performance for the object recognition task. Deep networks are hard to train due to the notorious vanishing gradient problem \cite{krizhevsky2012alexnet}. Since the gradient is back-propagated to earlier layers, repeated multiplication may make the gradient infinitively small. As a result, when network becomes deeper, its performance may degrade rapidly. ResNet is based on the idea of skip connections which ensures optimal hyperparameter values such as number of layers and overcomes the vanishing gradient problem. \\
\textbf{Inception-ResNet} \cite{szegedy2017inception} as the name suggests is a hybrid architecture of Inception and ResNet. Essentially, the pooling operations in the inception modules have been replaced with residual connections. However these changes are only to the Inception blocks and the reduction blocks remain intact. There are two variants of Inception-ResNet network, denoted as v1 and v2, whose only difference lies in the hyperparameter settings. Inception-ResNet has been shown to achieve superior performance with a lower number of training epochs. \\
\textbf{EfficientNet} \cite{tan2019efficientnet} is one of the most recent architectures based on a novel scaling method that uses compound coefficient to scale networks. Unlike conventional approaches that arbitrarily scale network dimensions, such as width, depth and resolution, this method uniformly scales each dimension with a fixed set of scaling coefficients. Based on this scaling method and advancements in NAS lead to the development of a family of models known as EfficientNet which offers a ten fold efficiency gain when compared to the state of the art for ImageNet classification \cite{russakovsky2015imagenet}. \\

\subsection{Simplified CNN Architectures} \label{ssec:simCNNArch}
Our experimental approach systematically investigated variations in architectural configurations of each reference architecture. Performance is measured using the same evaluation parameters set out in Section \ref{sec:exp}.

For InceptionV2 we consider three different variants and four different sub-variants. Each of the three variants consists only of the inception modules illustrated in \fig \ref{fig:inception_blocks} (Modules A-C) to facilitate separate evaluation. Since the primary goal of this work is to develop a simplified CNN architecture, we restrict the maximum number of inception modules to six in each of the three major variants. The various network variants evaluated for each of the major variants are denoted as follows: 
\begin{itemize}
    \item {\it A3-A6 -} A variant consisting of only Module-A components, where $A$ contains $n$ modules for $n=\{3..6\}$.
    \item {\it B3-B6 -} consisting of only Module-B components which have asymmetric factorization of $7 \times7$ convolutions, where $B$ contains $n$ modules for $n=\{3..6\}$.
    \item {\it C3-C6 -} consisting of only Module-C components which are broadened and concatenated with $n$ modules for $n=\{3..6\}$.
\end{itemize}

InceptionV3 is architecturally modified into 12 different variants with the naming convention $InceptionV3_{v01-12}$. Each of these variants use different combinations of inception modules (Figures \ref{fig:inception_blocks}/\ref{fig:grid_size_reduction}). The first six variants have the same number of filters as mentioned in the original work \cite{szegedy2016rethinking} and the latter six variants have reduced number of filters according to Eq. \ref{eq:filter}. We restrict the number filters in each layer less than 100, and secondly, the number of filters is a multiple of the original number of filters as in the original work \cite{szegedy2016rethinking}. If the original number of filter is $M$ in a layer, then the reduced number of filter ($M_r$) is calculated as follow: 

\begin{equation} \label{eq:filter}
M_r = \left\{\begin{matrix}
\mathlarger{\mathlarger{\mathlarger{\mathlarger{\lfloor}}}} \enspace \frac{M}{2^{\lceil \log_2 \frac{M}{100} \rceil}} \mathlarger{\mathlarger{\mathlarger{\mathlarger {\rceil}}}} \enspace \qquad  M > 100 
\\ \\
M \quad \qquad  \qquad \qquad otherwise
\end{matrix}\right.
\end{equation}


Each variation of InceptionV3, which is a combination of Module A/B/C, grid size reduction (GR) of Module - A/B and reduced number of filters applied (according to Eq. \ref{eq:filter}) and connected one upon the other, is presented in Table \ref{inceptionv3variants}. Similar to InceptionV3, InceptionV4 is modified into 12 different variants. 
As with InceptionV3, the first six variants ($v01-06$) consist of the network variants with the same number of filters as mentioned in the original work for InceptionV4 \cite{szegedy2017inception} and the latter six variants ($v07-12$) consist of both reduced number of filters with a modified stem to reduce the computational complexity. The InceptionV4 variants also follow the similar naming convention as InceptionV3 with variants being named as $InceptionV4_{v01-12}$. Each variants of InceptionV4 is presented in Table \ref{inceptionv4variants}.

\begin{table*}[ht]
\centering
    \caption{InceptionV3 variants with different components.}
\begin{tabular}{lcccccc}
\hline
Architecture & Module-A & GR-A & Module-B & GR-B & Module-C & Reduced filter  \\ \hline
$InceptionV3_{v01}$ & \greencheck&\greencheck & \greencheck & \greencheck & \greencheck \\ 
$InceptionV3_{v02}$ & \greencheck &  & \greencheck &\greencheck & \greencheck \\ 
$InceptionV3_{v03}$ & \greencheck && \greencheck & &\greencheck \\ 
$InceptionV3_{v04}$ &  & \greencheck&\greencheck & \greencheck & \greencheck \\ 
$InceptionV3_{v05}$ & &\greencheck & \greencheck & \greencheck&  \\ 
$InceptionV3_{v06}$ & & \greencheck& \greencheck & \greencheck & \\ \hdashline
$InceptionV3_{v07}$  & &\greencheck & \greencheck & \greencheck&  & \greencheck \\ 
$InceptionV3_{v08}$ & \greencheck & \greencheck & \greencheck &\greencheck & &\greencheck \\ 
$InceptionV3_{v09}$  & \greencheck && \greencheck & &\greencheck &\greencheck \\ 
$InceptionV3_{v10}$  &  & \greencheck&\greencheck & \greencheck & \greencheck &\greencheck \\ 
$InceptionV3_{v11}$  & \greencheck &  & \greencheck &\greencheck & \greencheck &\greencheck  \\ 
$InceptionV3_{v12}$  & \greencheck&\greencheck & \greencheck & \greencheck & & \greencheck \\ \hline
        
\end{tabular}
    \label{inceptionv3variants}
\end{table*}

\begin{table*}[ht]
    \centering
        \caption{InceptionV4 variants with different components.}
    \begin{tabular}{lcccccc}
    \hline
Architecture & Module-A & GR-A & Module-B & GR-B & Module-C & Reduced filter  \\ \hline
$InceptionV4_{v01}$ & \greencheck&\greencheck & \greencheck & \greencheck & \greencheck & \\ 
$InceptionV4_{v02}$ & \greencheck & \greencheck & \greencheck & & \greencheck & \\ 
$InceptionV4_{v03}$  & \greencheck & & \greencheck & \greencheck&\greencheck &\\ 
$InceptionV4_{v04}$  & \greencheck & \greencheck& & \greencheck & \greencheck &\\ 
$InceptionV4_{v05}$  & \greencheck& & \greencheck & & \greencheck &\\
$InceptionV4_{v06}$  & & \greencheck& \greencheck& \greencheck & \greencheck & \\ \hdashline
$InceptionV4_{v07}$ & \greencheck&\greencheck & \greencheck & \greencheck & \greencheck & \greencheck \\ 
$InceptionV4_{v08}$ & \greencheck & \greencheck & \greencheck &\greencheck & & \greencheck\\ 
$InceptionV4_{v09}$  & \greencheck & & \greencheck & \greencheck&\greencheck & \greencheck\\ 
$InceptionV4_{v10}$  & & \greencheck&\greencheck & \greencheck & \greencheck & \greencheck\\ 
$InceptionV4_{v11}$  & \greencheck& & \greencheck &  \greencheck& & \greencheck\\ 
$InceptionV4_{v12}$  & \greencheck& \greencheck& \greencheck & \greencheck & \greencheck & \greencheck \\ \hline
        
    \end{tabular}
    \label{inceptionv4variants}
\end{table*}

ResNet has been evaluated as it is with varying depths. The four ResNet models are ResNet-\{{\it 18,34,50,101}\}. Inception-ResNet v1 and v2 have been evaluated with no modifications. Three different variations of EfficientNet-\{{\it B0,B1,B2}\}, defined as in the original work \cite{tan2019efficientnet}, has been evaluated.

Based on an exhaustive set of experimentation over the full set of variants outlined, under the conditions outlined in Section \ref{sec:proposal}, we experimentally identify and propose the following two maximally reduced complexity performing architectural variants targeted towards the fire detection and localization task.

\begin{figure}[tb]
\centering
\includegraphics[width=\linewidth]{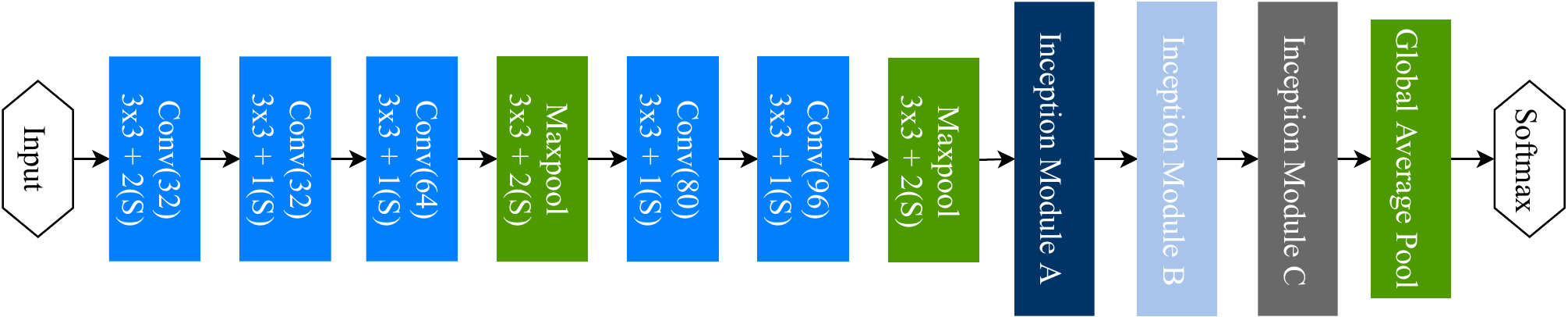}
\vspace{-0.6cm}
\caption{Reduced complexity architecture for InceptionV3-OnFire optimized for fire detection.}
\label{fig:inceptionv3onfire}
\end{figure}

\textbf{InceptionV3-OnFire} is inspired by the performance of the  $InceptionV3_{v09}$ variant. One of each Inception Module - A, B, and C are connected to develop the InceptionV3-OnFire architecture, as illustrated in \fig \ref{fig:inceptionv3onfire}. To reduce the complexity, the number of filters in each layer are restricted as according to the Eq. \ref{eq:filter}.

\begin{figure}[tb]
\centering
\includegraphics[width=\linewidth]{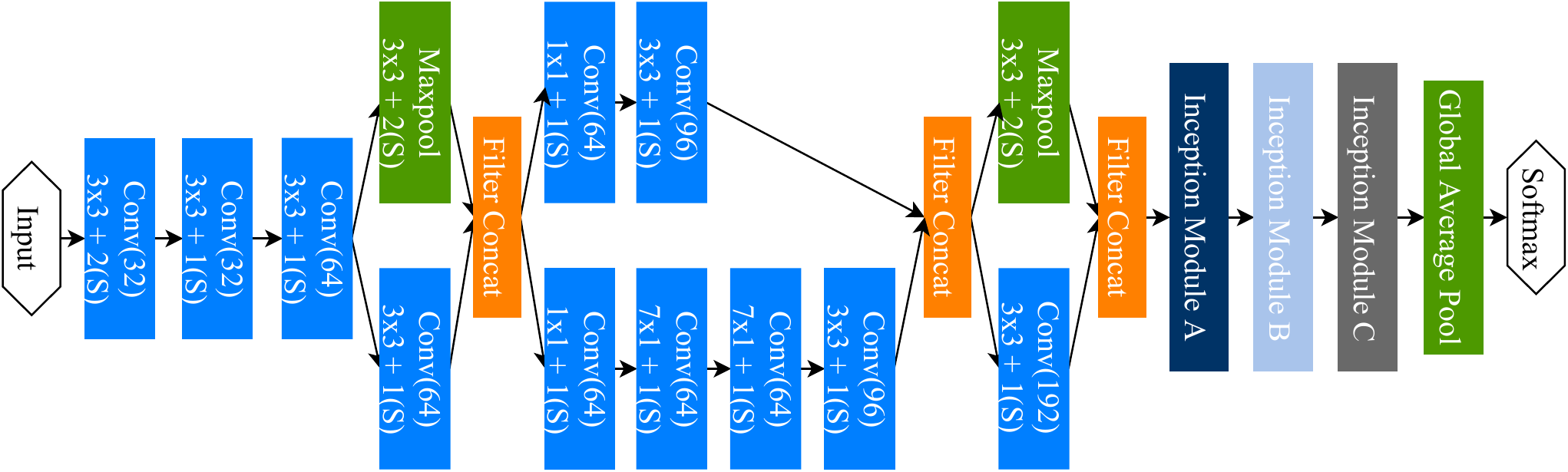}
\vspace{-0.6cm}
\caption{Reduced complexity architecture for InceptionV4-OnFire optimized for fire detection.}
\label{fig:inceptionv4onfire}
\end{figure}

\textbf{InceptionV4-OnFire} is a three layered version of InceptionV4 which is based on the $InceptionV4_{v05}$ variant containing one each of the three inception Module- A, B and C (\fig \ref{fig:inceptionv4onfire}). The grid size reduction module is removed. Each of the inception modules followed the same definition as that of the original work. A dropout of 0.4 is applied at the end of the network to prevent the model from over-fitting. The same stem as in the original InceptionV4 architecture is used as illustrated in \fig \ref{fig:inceptionv4onfire}.  

Overall, the governing intuition based on these variants is that, a combination of all the three inception modules (\fig \ref{fig:inception_blocks}) will perform better as the architecture is equipped with both depth and width to optimally learn how to detect and localize fire. The grid reduction modules are mainly used to shrink the height and width of the image in a more optimal fashion, although it eventually leads to information loss. 
\subsection{Superpixel Localization} \label{ssec:superpixel}
Further expanding this work, we adopt the use of image over-segmentation based fire localization, contrary to the earlier works \cite{Phillips02fire} \cite{CELIK2009fire}\cite{yuan2012smoke}, which rely on colour based fire localization. Superpixel based approaches over-segment the image into perceptually meaningful regions which are similar in colour and texture. Specifically we incorporate the Simple Linear Iterative Clustering (SLIC) \cite{achanta2012slic} over-segmentation approach, which performs iterative clustering in a similar manner to {\it k-means} to reduced spatial dimensions, where the image is segmented into approximately equally sized superpixels (\fig \ref{fig:examples_fire_local}). Each over-segmented/superpixel image region is subsequently classified using proposed InceptionV3-OnFire/InceptionV4-OnFire architecture formulated as a \{{\it fire, no-fire}\}, for fire detection task. To boost the performance, we additionally use the network weight initialization via transfer learning from the primary full frame binary fire detection task.

    \section{Experimental Setup} \label{sec:exp}
We address the problem of full frame binary fire detection ({\it i.e. is there fire present in the image as a whole - yes/no?}) as well as fire localization ({\it i.e location of the fire in the image?}). All networks are trained using Nvidia GeForce GTX 1080Ti GPU via TensorFlow (1.13.1 + TFLearn 0.3.2).  
The network variants are tested with different optimizers such as stochastic gradient descent (SGD) with momentum and Root Mean Square Propagation (RMSProp) along with normalization techniques such as local response normalization and batch normalization. 
The training is performed with categorical cross entropy loss, for 30 epochs and a learning rate of 0.001.

\subsection{Full-frame Binary Fire Detection} \label{exp:binary}
For the binary fire detection problem, network training and testing are performed on the dataset created in the work \cite{dunnings2018experimentally} which consists of 23,408 images. This dataset is split (80:20 split) into two portions for training and validation. An additional set of 2,931 images was used for cross validation.

\subsection{Superpixel Localization Setup} \label{exp:superpixel}
To evaluate within the context of in-frame localization, we use the dataset created in the work \cite{dunnings2018experimentally}. The network architectures are trained on a total of 8,635 fire superpixel images, and 10,000 non-fire superpixel images with a test set of 3,000 images containing 1,500 fire and 1,500 non-fire examples.
These images are further pre-processed to centre the superpixel region to make it  location independent and padded to a size of $224 \times 224$. 
 
\section{Evaluation} \label{sec:eval}
For statistically comparing different architectures we consider the true positive rate (TPR), false positive rate (FPR) along with F-score (F), Precision (P) and Accuracy (A), Complexity (number of parameters in millions, C), the ratio between accuracy and number of parameters (A:C) and achievable frames per second (fps) throughput.

The results of full-frame binary fire detection are presented in Table \ref{Tab:binary}. We present only the best performing variants of the reference architectures (Sec. \ref{ssec:refArch}) with results shown in the Table \ref{Tab:binary} (middle). From the results, we can observe that our proposed variant of InceptionV4, InceptionV4-OnFire, offers the best performance (Table \ref{Tab:binary}, lower) in terms of accuracy, TPR (A: 0.96, TPR: 0.95) compared to other architectures. Both of our proposed architectures, InceptionV3-OnFire and InceptionV4-OnFire, achieve the lowest false positive rate (FPR: 0.07/0.04, Table \ref{Tab:binary}, lower), compared to previous work of FireNet \cite{dunnings2018experimentally} with (FRP: 0.09 FPR, Table \ref{Tab:binary}, upper).  
The reduced complexity, InceptionV3 variant performs just marginally worse when compared to the InceptionV4 variant but still outperforms InceptionV1-OnFire \cite{dunnings2018experimentally} in terms of accuracy, false positive rate and the accuracy is to number of parameters ratio (Table \ref{Tab:binary}). 

\begin{table}[ht]
\caption{Statistical performance for full-frame binary fire detection. Upper: Prior work. Middle: Reference architectures. Lower: Our approaches.}
\begin{tabular}{lllllll}
\hline
Architecture &TPR  & FPR &F1 &P&A \\ \hline

\small FireNet \cite{dunnings2018experimentally} & 0.92 & 0.09  &0.93  & 0.93 & 0.92  \\  
\small InceptionV1-OnFire \cite{dunnings2018experimentally} & 0.96 & 0.10  &0.94  & 0.93 & 0.93  \\ \hline

\small InceptionV2-B6 & 0.97 & 0.09 & 0.95 & 0.94 & 0.95 \\ 
\small ResNet-18 & 0.92 & 0.05 & 0.94 & 0.96 & 0.93 \\ 
\small Inception-ResNet-v1 & 0.84& {\bf 0.03} & 0.90 & 0.97 & 0.89 \\ 
\small EfficientNet-B0 & 0.94 & 0.16 & 0.91 & 0.89 & 0.90 \\ \hline

\small InceptionV3-OnFire & {\bf 0.95} & 0.07  &0.95  & 0.95& 0.94 \\
\small {\bf InceptionV4-OnFire} & {\bf 0.95} & 0.04  & {\bf 0.96}  & {\bf 0.97} & \textbf{0.96}  \\  \hline
\end{tabular}

\label{Tab:binary}
\end{table}

\begin{table}[ht]
\caption{Computational Efficiency for full-frame binary fire detection.}
\begin{tabular}{lllll}
\hline
Architecture &C  & A(\%) &A:C &fps \\ \hline
\small FireNet \cite{dunnings2018experimentally} & 68.3 & 91.5  &1.3  & \textbf{17}  \\  
\small InceptionV1-OnFire \cite{dunnings2018experimentally} & 1.2 & 93.4  &77.9  & 8.4   \\ \hline

\small \textbf{InceptionV3-OnFire} & 0.96 & 94.4  &\textbf{98.09}  & 13.8 \\
\small InceptionV4-OnFire  & 7.18 & 95.6  &13.37   & 12  \\ \hline 
\end{tabular}
\label{Tab:comput}
\end{table}

Conversely, we find that the InceptionV3 variant marginally outperforms the InceptionV4 variant in terms computational efficiency (A:C, fps in Table \ref{Tab:comput}-lower). Although the number of parameters is reduced to 0.96 million in InceptionV3-OnFire compared to 68.3/1.2 million in FireNet/InceptionV1OnFire \cite{dunnings2018experimentally}, the run-time throughput is still higher for FireNet (Table \ref{Tab:comput}). Whilst FireNet \cite{dunnings2018experimentally} provides a maximal throughput of 17 fps, it is notable that InceptionV3-OnFire provides the maximal accuracy to complexity ratio. 

\begin{table}[ht]
\caption{Localization Results - within frame superpixel approach.}
\begin{tabular}{llllll}
\hline
Architecture & TPR& FPR& F& P & A\\ \hline
\small InceptionV1-OnFire \cite{dunnings2018experimentally} & 0.92 & 0.17  &0.9  & 0.88&0.89   \\ \hline
\small InceptionV3-OnFire & 0.94 & 0.07  &0.94  & 0.93 &\textbf{0.94}   \\  
\small {\bf InceptionV4-OnFire} & {\bf 0.94} & \textbf{0.06}  & {\bf 0.94}   & {\bf 0.94} &\textbf{0.94} \\ \hline 
\end{tabular}
\label{Tab:localization}
\end{table}

The results for superpixel based fire localization are presented in Table \ref{Tab:localization} where we can see that InceptionV4-OnFire marginally outperforms InceptionV3-OnFire in terms of a lower FPR with equal overall accuracy representing a 5\% increase in performance over prior work in the field (InceptionV1-OnFire \cite{dunnings2018experimentally}). 

\begin{figure}[tb]
\centering
\includegraphics[width=\linewidth]{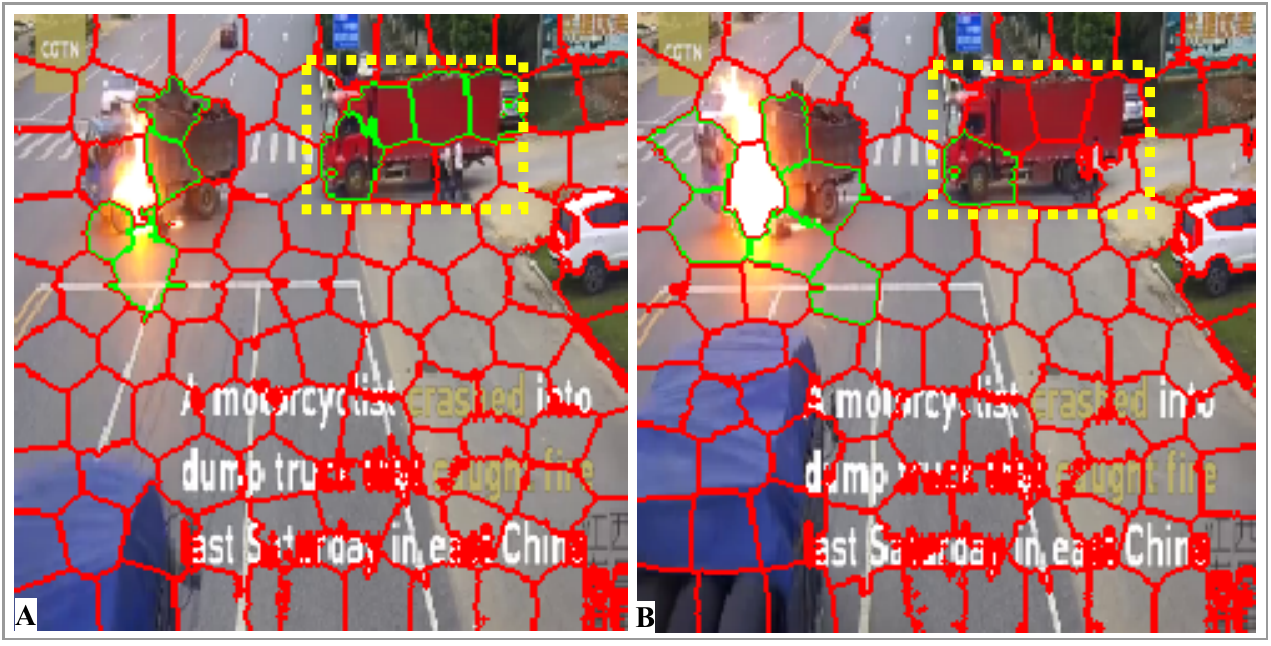}
\caption{Comparison of results in yellow dashed box without transfer learning (A) and with transfer learning (B), where fire = green, no-fire = red.}
\label{fig:superpixel_transer_learning}
\vspace{-0.4cm}
\end{figure}

Qualitative examples of this localization (InceptionV4-OnFire), including the canonoical challenge of red coloured non-fire regions, are illustrated in Figure \ref{fig:superpixel_transer_learning}. From this figure, we see the positive performance impact of transfer learning from the initial full-frame binary fire detection into this fire localization task. The region inside yellow dashed box in the \fig \ref{fig:superpixel_transer_learning}A is falsely detected is {\it fire}, however, with transfer learning the same region is correctly detected as {\it no-fire} in \fig \ref{fig:superpixel_transer_learning}B. Transfer learning significantly reduces this type of FP occurrence by approximately 10\% (FPR: 0.06 from previously 0.17, Table \ref{Tab:localization}).

From Tables \ref{Tab:comput} and \ref{Tab:binary} we can see that InceptionV4-OnFire/InceptionV3-OnFire offer slightly lesser computational performance in terms of frame-rate than prior work (FireNet, \cite{dunnings2018experimentally}) but significantly improved detection performance.
    \section{Conclusion}
Our proposed reduced complexity CNN architecture (InceptionV4-OnFire), which is experimentally defined from leading CNN architectures, achieve maximal 0.96 accuracy for binary full-frame fire detection task. We significantly reduce the false positive rate as low as 0.04 outperforming the prior state-of-the-art approach of FireNet \cite{dunnings2018experimentally}. We also manage reduce the number of parameters of InceptionV3-OnFire by 0.24 million compared to architectures in \cite{dunnings2018experimentally}. Furthermore, for superpixel based fire detection, we notably reduce the false positive rate to 0.06 by employing a transfer learning strategy. Overall, this work presents robust and reduced complexity architectures for full-frame/superpixel fire detection task enabled by extended, exhaustive experimental evaluation.

\small{
\bibliographystyle{IEEEtran}
\bibliography{refs,icip-2016-refs}
}
\end{document}